\documentclass[10pt,twocolumn,letterpaper]{article}

\usepackage{iccv}
\usepackage{times}
\usepackage{epsfig}
\usepackage{graphicx}
\usepackage{subcaption}
\usepackage{amsmath}
\usepackage{amssymb}
\usepackage{bm}
\usepackage{textcomp}
\usepackage{enumitem}

\usepackage[breaklinks=true,bookmarks=false]{hyperref}

\iccvfinalcopy % *** Uncomment this line for the final submission

\def\iccvPaperID{****} % *** Enter the ICCV Paper ID here

\begin{document}

%%%%%%%%% TITLE
\title{Learning to Collocate Neural Modules for Image Captioning}

\author{Xu Yang, Hanwang Zhang, Jianfei Cai\\
School of Computer Science and Engineering,\\
Nanyang Technological University,\\
\tt\small{ s170018@e.ntu.edu.sg,\{hanwangzhang@,ASJFCai@\}ntu.edu.sg}
}

\maketitle
%\thispagestyle{empty}

%%%%%%%%% ABSTRACT
\begin{abstract}
   We do not speak word by word from scratch; our brain quickly structures a pattern like \textsc{sth do sth at someplace} and then fill in the detailed descriptions. To render existing encoder-decoder image captioners such human-like reasoning, we propose a novel framework: learning to Collocate Neural Modules (CNM), to generate the ``inner pattern'' connecting visual encoder and language decoder. Unlike the widely-used neural module networks in visual Q\&A, where the language (\ie, question) is fully observable, CNM for captioning is more challenging as the language is being generated and thus is partially observable. To this end, we make the following technical contributions for CNM training: 1) compact module design --- one for function words and three for visual content words (\eg, noun, adjective, and verb), 2) soft module fusion and multi-step module execution, robustifying the visual reasoning in partial observation, 3) a linguistic loss for module controller being faithful to part-of-speech collocations (\eg, adjective is before noun). Extensive experiments on the challenging MS-COCO image captioning benchmark validate the effectiveness of our CNM image captioner. In particular, CNM achieves a new state-of-the-art 127.9 CIDEr-D on Karpathy split and a single-model 126.0 c40 on the official server. CNM is also robust to few training samples, \eg, by training only one sentence per image, CNM can halve the performance loss compared to a strong baseline. 
\end{abstract}

%%%%%%%%% BODY TEXT
\section{Introduction}
\label{sec:intro}

\begin{figure}[t]
\centering
    \begin{subfigure}[t]{.9\linewidth}
        \includegraphics[width=1\linewidth,trim = 5mm 5mm 5mm 5mm,clip]{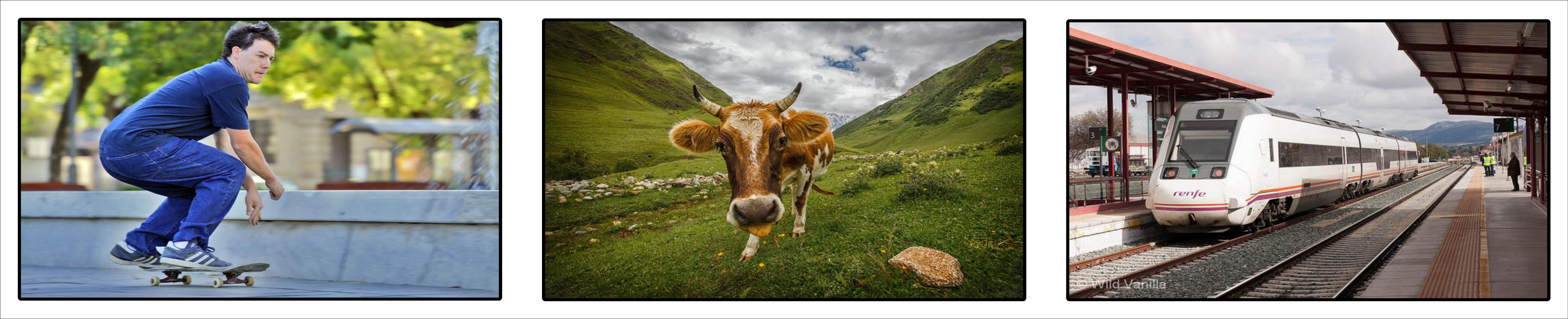}
        \caption{Three diverse images. }
        \label{fig:introa}
    \end{subfigure}
    \begin{subfigure}[t]{.9\linewidth}
        \includegraphics[width=1\linewidth,trim = 5mm 5mm 5mm 5mm,clip]{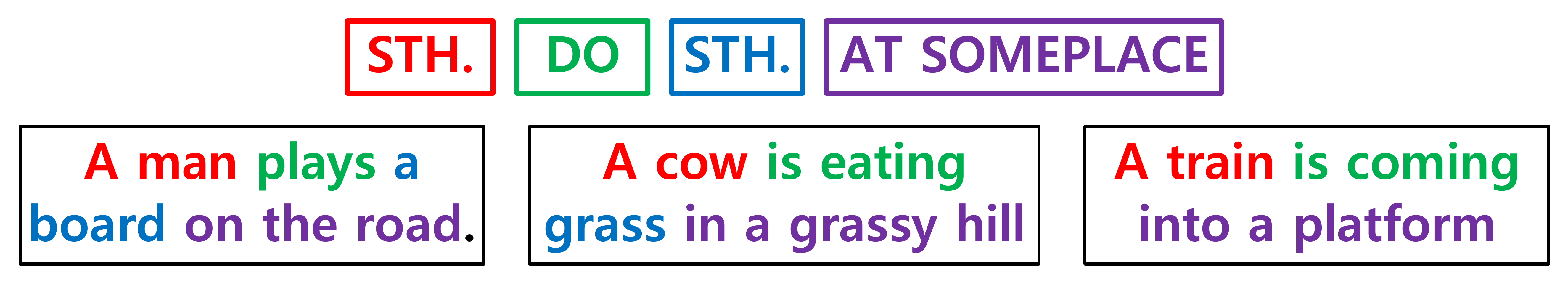}
        \caption{Three captions with the same sentence pattern.}
        \label{fig:introb}
    \end{subfigure}
    \begin{subfigure}[t]{.9\linewidth}
        \includegraphics[width=1\linewidth,trim = 5mm 5mm 5mm 5mm,clip]{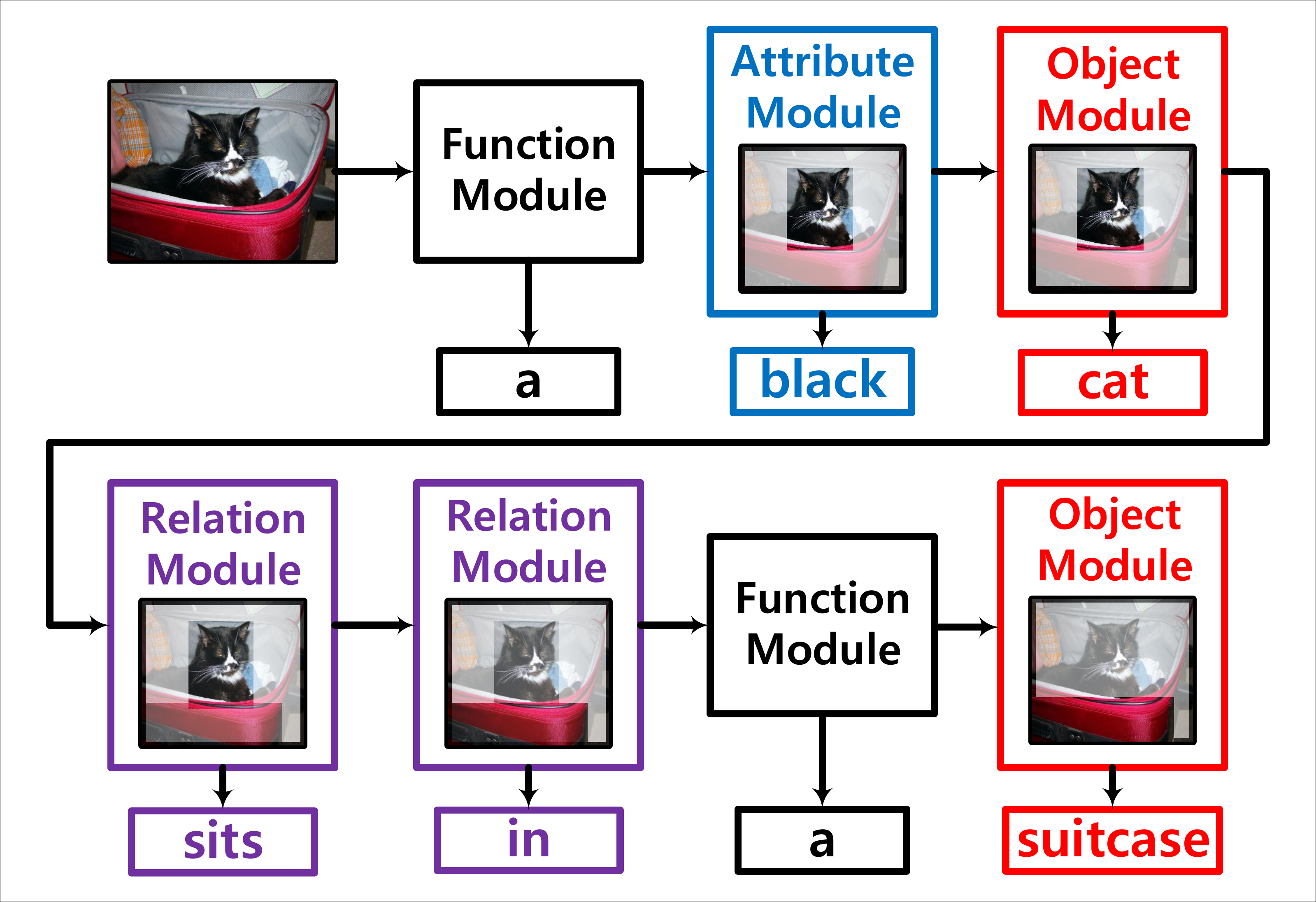}
        \caption{The caption generation process of CNM.}
        \label{fig:introc}
    \end{subfigure}
     \caption{The motivation of the proposed learning to Collocate Neural Modules (CNM) for image captioning: neural module collocation imitates the inductive bias --- sentence pattern, which regularizes the diverse training effectively.}
     \vspace{-0.2in}
\end{figure}

\begin{figure*}[t]
\centering
\includegraphics[width=1\linewidth,trim = 5mm 5mm 5mm 5mm,clip]{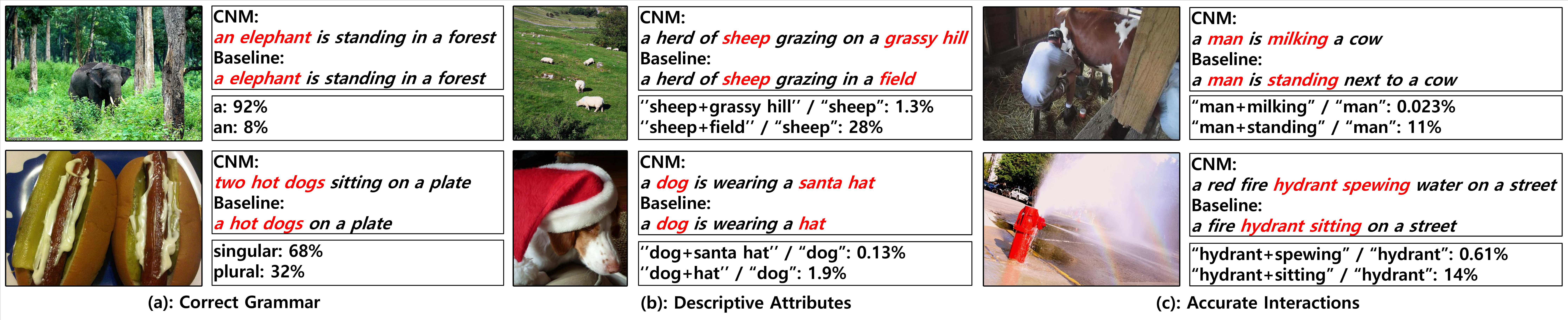}
  \caption{By comparing our CNM with a non-module baseline (an upgraded version of Up-Down~\cite{anderson2018bottom}), we have three interesting findings in tackling the dataset bias: (a) more accurate grammar. \% denotes the frequency of a certain pattern in MS-COCO, (b) more descriptive attributes, and (c) more accurate object interactions. The ratio ./. denotes the percentage of co-occurrence, \eg, ``sheep+field''/``sheep'' = 28\% means that ``sheep'' and ``field'' contributes the 28\% occurrences of ``sheep''. We can see that CNM outperforms the baseline even with highly biased training samples.}
  \vspace{-0.2in}
\label{fig:intro2}
\end{figure*}

Let's describe the three images in Figure~\ref{fig:introa}. Most of you will speak sentences varying vastly from image to image. In fact, the ability of using diverse language to describe the colorful visual world is a gift to humans, but a formidable challenge to machines. Although recent advances in visual representation learning~\cite{he2016deep, ren2015faster, he2017mask} and language modeling~\cite{hochreiter1997long, vaswani2017attention} demonstrate the impressive power of modeling the diversity in their respective modalities, it is still far from being resolved to establish a robust cross-modal connection between them. Indeed, image captioning is not the only model that can easily exploit the dataset bias to captioning even without looking at the image, almost all existing models for vision-language tasks such as visual Q\&A~\cite{johnson2017clevr,goyal2017making,shi2018explainable} have been spotted mode collapse to certain dataset idiosyncrasies, failed to reproduce the diversity of our world --- the more complex the task is, the more severe the collapse will be, such as image paragraph generation~\cite{krause2017hierarchical} and visual dialog~\cite{das2017visual}. For example, in MS-COCO~\cite{lin2014microsoft} training set, as the co-occurrence chance of ``man'' and ``standing'' is 11\% large, a state-of-the-art captioner~\cite{anderson2018bottom} is very likely to generation ``man standing'', regardless of their actual relationships such as ``milking'', which is 0.023\% rare. We will discuss more biased examples in Figure~\ref{fig:intro2} later.

Alas, unlike a visual concept in ImageNet which has 650 training images on average~\cite{deng2009imagenet}, a specific sentence in MS-COCO has \emph{only one single} image~\cite{lin2014microsoft}, which is extremely scarce in the conventional view of supervised training. However, it is more than enough for us humans --- anyone with normal vision (analogous to pre-trained CNN encoder) and language skills (analogous to pre-trained language decoder) does NOT need any training samples to perform captioning. Therefore, even though substantial progress has been made in the past 5 years since Show\&Tell~\cite{vinyals2015show}, there is still a crucial step missing between vision and language in modern image captioners~\cite{anderson2018bottom,lu2017knowing,lu2018neural}. To see this, given a sentence pattern in Figure~\ref{fig:introb}, your descriptions for the three images in Figure~\ref{fig:introa} should be much more constrained. In fact, studies in cognitive science~\cite{hale2018finding,slevc2011saying} show that do us humans not speak an entire sentence word by word from scratch; instead, we compose a pattern first, then fill in the pattern with concepts, and we repeat this process until the whole sentence is finished. Thus, structuring such patterns is what our human ``captioning system'' practices every day, and should machines do so. Fortunately, as we expected, for the sentence pattern in Figure~\ref{fig:introb}, besides those three captions, we have thousands more in MS-COCO. 

In this paper, we propose learning to Collocate Neural Modules (CNM) to fill the missing gap in image captioning, where the module collocation imitates the sentence pattern in language generation. As shown in Figure~\ref{fig:introc}, CNM first uses the \textsc{function} module for generating function word ``a'', and then chooses the \textsc{attribute} module to describe the adjectives like ``black'' of the ``cat'', which will be generated by the \textsc{object} module for nouns, followed by \textsc{relation} module for verbs or relationships like ``sits in''. Therefore, the key of CNM is to learn a dynamic structure that is an inductive bias being faithful to language collocations. 

Though using neural module networks is not new in vision-language tasks such as VQA~\cite{andreas2016neural}, where the question is parsed into a module structure like \textsc{Color}(\textsc{Find}(`chair')) for ``What color is the chair?''; for image captioning, the case is more challenging as only partially observed sentences are available during captioning, and the module structure by parsing is no longer applicable. To this end, we develop the following techniques for effective and robust CNM training. 1) Inspired by the policy network design in partially observed environment reinforcement learning~\cite{foerster2016learning}, at each generation time step, the output of the four modules will be the fused according to their soft attention, which is based on the current generation context. 2) We adopt multi-step reasoning, \ie, stacking neural modules~\cite{hu2018explainable}. These two methods stabilize the CNM training greatly. 3) To further introduce expert knowledge, we impose a linguistic loss for the module soft attention, which should be faithful to part-of-speech collocations, \eg, \textsc{attribute} module should generate words that are ADJ. 

Before we delve into the technical details in Section~\ref{sec:cnm}, we would like to showcase the power of CNM in tackling the dataset bias in Figure~\ref{fig:intro2}. Compared to a strong non-module baseline~\cite{anderson2018bottom}, the observed benefits of CNM include: 1) more accurate grammar like less `a/an' error and `singular/plural' error (Figure~\ref{fig:intro2}a), thanks to the joint reasoning of \textsc{function} and \textsc{object} module, 2) more descriptive attributes (Figure~\ref{fig:intro2}b) due to \textsc{attribute} module, and 3) more accurate interactions (Figure~\ref{fig:intro2}c) due to \textsc{relation} module. Moreover, we find that when only 1 training sentence of each image is provided, our CNM will suffer less performance deterioration compared with the strong baseline. Extensive discussions and human evaluations are offered in Section~\ref{sec:experiment}, where we validate the effectiveness of  CNM on the challenging MS-COCO image captioning benchmark. Overall, we achieve 127.9 CIDEr-D score on Karpathy split and a single model 126.0 c40 on the official server. 

Our contributions are summarized as follows:
\begin{itemize}[leftmargin=.1in]
\item Our CNM is the first module networks for image captioning. This enriches the spectrum of using neural modules for vision-language tasks.

\item We develop several techniques for effective module collocation training in partially observed sentences.

\item Experiment results show that significant improvement can be made by using neural modules. CNM is a generic framework that supports potential improvement like more principled module and controller designs.
\end{itemize}

\section{Related Work}
\textbf{Image Captioning}.
Most early image captioners are template-based models that they first structure sentence patterns and then fill the words into these fixed patterns~\cite{kulkarni2011baby,kuznetsova2012collective,mitchell2012midge}.
However, since the functions used for generating templates and for generating words are not jointly trained, the performances are limited. Compared with them, modern image captioners which achieve superior performances are attention based encoder-decoder methods~\cite{xu2015show,vinyals2015show,rennie2017self,liu2018context,lu2017knowing,anderson2018bottom,yao2017boosting,lu2018neural}. However, unlike the template based models, most of the encoder-decoder based models generate word one by one without structure. Our CNM makes full use of the advantages of both template and encoder-decoder based image captioners which can generate captions by structuring patterns and  end-to-end training. In particular, from the perspective of module network, several recent works can be reduced to a special case of our CNM. For example, Up-Down~\cite{anderson2018bottom} only adopts \textsc{object} module, and NBT~\cite{lu2018neural} only uses \textsc{object} and \textsc{function} modules while they treat all the non-object words as function words.

\textbf{Neural Module Networks}. Recently, the idea of decomposing the network into neural modules is popular in some vision-language tasks such as VQA~\cite{andreas2016neural,hu2017learning}, visual grounding~\cite{liu2018explainability,yu2018mattnet}, and visual reasoning~\cite{shi2018explainable}. In these tasks, high-quality module layout can be obtained by parsing the provided sentences like questions in VQA. Yet in image captioning, only partially observed sentences are available and the module structure by parsing is not applicable anymore. For addressing such a challenge, we propose to dynamically collocate neural modules on-the-fly during sentence generation.

\section{Learning to Collocate Neural Modules}
\label{sec:cnm}
\begin{figure}[t]
\centering
\includegraphics[width=1\linewidth,trim = 5mm 5mm 5mm 5mm,clip]{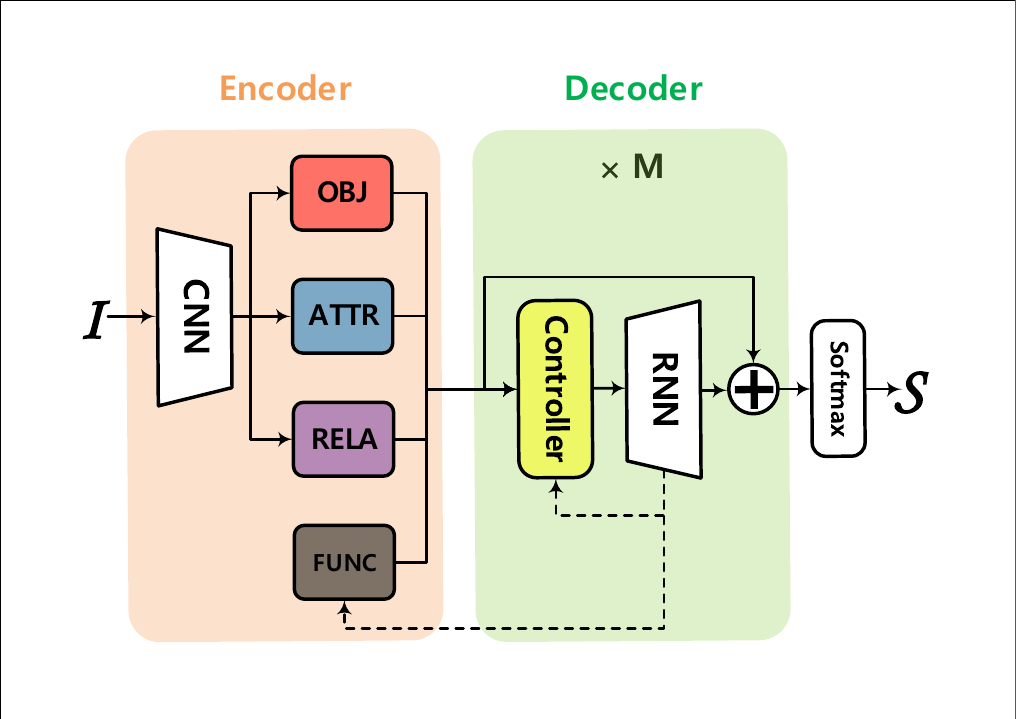}
   \caption{The encoder-decoder pipeline of our learning to Collocate Neural Modules (CNM) image captioner. The dash lines from RNN to \textsc{function} module and the module controller mean that both of these sub-networks require the contextual knowledge of partially observed sentences.}
\label{fig:pip}
\end{figure}

Figure~\ref{fig:pip} shows the encoder-decoder structure of our learning to Collocate Neural Modules (CNM) model. The encoder contains a CNN and four neural modules to generate features for language decoding (cf. Section~\ref{subsec:modules}). Our decoder has a module controller that softly fuses these features into a single feature for further language decoding by the followed RNN (cf. Section~\ref{sec:soft_fuse}). Note that a linguistic loss is imposed for making the module controller more faithful to part-of-speech collocations (cf. Section~\ref{sec:ling_loss}). Besides the language generation, the RNN would also output the accumulated context of the partially observed sentence as the input to \textsc{function} module and controller for linguistic information, which is helpful for these grammar-related modules. For multi-step reasoning, the entire decoder of CNM will repeat this soft fusion and language decoding $M$ times (cf. Section~\ref{sec:multi_res}). The residual connections are also implemented for directly transferring knowledge from lower layers to higher ones. 

\subsection{Neural Modules}
\label{subsec:modules}
Four neural modules are designed for predicting the orthogonal knowledge from the image, \eg, \textsc{object} module focuses on the object categories while \textsc{attribute} module focuses on the visual attributes. In this way, the caption generation can be disentangled from dataset bias at the word-level, \ie, the words are generated from the visual knowledge from each module, not merely from the language context which is more likely overfitted to dataset bias. For example, the more accurate description ``bird-perch-tree'' will be reduced to ``bird-fly'' without using \textsc{relation} module, due to the high co-occurrence of ``bird'' and ``fly'' in the dataset. Now, we detail each of the modules.

\noindent\textbf{\textsc{Object} Module.} 
It is designed to transform the CNN features to a feature set $\mathcal{V}_O$ containing the knowledge on object categories, \ie, the feature set $\mathcal{V}_O$ facilitates the prediction of nouns like ``person'' or ``dog''. The input of this module is $\mathcal{R}_O$, which is an $N \times d_r$ feature set of $N$ RoI features extracted by a ResNet-101 Faster R-CNN~\cite{ren2015faster}. This ResNet is pre-trained on object detection task by using the object annotations of VG dataset~\cite{krishna2017visual}. Formally, this module can be formulated as:
\begin{equation} \label{equ:obj_mod}
\small
\begin{aligned}
 \textbf{Input:} \quad  &\mathcal{R}_O, \\
 \textbf{Output:} \quad  &\mathcal{V}_O = \text{LeakyReLU}(\text{FC}(\mathcal{R}_O)), \\
\end{aligned}
\end{equation}
where $\mathcal{V}_O$ is the $N \times d_v$ output feature set.

\noindent\textbf{\textsc{Attribute} Module.} 
It is designed to transform the CNN features to a feature set $\mathcal{V}_A$ on attribute knowledge, for generating adjectives like ``black'' and ``dirty''. The input of this module is an $N \times d_r$ feature set extracted by a ResNet-101 Faster R-CNN, and the network used here is pre-trained on attribute classification task by using the attribute annotations of VG dataset. Formally, this module can be written as:
\begin{equation} \label{equ:attr_mod}
\small
\begin{aligned}
 \textbf{Input:} \quad  &\mathcal{R}_A, \\
 \textbf{Output:} \quad  &\mathcal{V}_A=\text{LeakyReLU}(\text{FC}(\mathcal{R}_A)), \\
\end{aligned}
\end{equation}
where $\mathcal{V}_A$ is the $N \times d_v$ feature set output from this module.

\noindent\textbf{\textsc{Relation} Module.} 
It transforms the CNN features to a feature set $\mathcal{V}_R$ representing potential interactions between two objects. This transferred feature set $\mathcal{V}_R$ would help to generate verbs like ``ride'', prepositions like ``on'', or quantifiers like ``two''. This module is built based on the multi-head self-attention mechanism~\cite{vaswani2017attention}, which automatically seeks the interactions among the input features. Here, we use $\mathcal{R}_O$ in Eq.~\eqref{equ:obj_mod} as the input because these kinds of features are widely applied as the input for successful relationship detection~\cite{zhang2017visual,zellers2018neural}. This module is formulated as:
\begin{equation} \label{equ:rela_mod}
\small
\begin{aligned}
 \textbf{Input:} \quad  &\mathcal{R}_O, \\
 \textbf{Multi-Head:} \quad & \mathcal{M}= \text{MultiHead}(\mathcal{R}_O), \\
 \textbf{Output:} \quad  &\mathcal{V}_R=\text{LeakyReLU}(\text{MLP}(\mathcal{M})), \\
\end{aligned}
\end{equation}
where MultiHead($\cdot$) means the multi-head self-attention mechanism, MLP($\cdot$) is a feed-forward network containing two fully connected layers with a ReLU activation layer in between~\cite{vaswani2017attention}, and $\mathcal{V}_R$ is the $N \times d_v$ feature set output from this module.
Specifically, we use the following steps to compute the multi-head self-attention. We first use scaled dot-product to compute $k$ self-attention head matrices as:
\begin{equation}
\small
    \textbf{head}_i=\text{Softmax}( \frac{\mathcal{R}_O\bm{W}_i^1(\mathcal{R}_O\bm{W}_i^2)^T}{\sqrt{d_k}} )\mathcal{R}_O\bm{W}_i^3,
\end{equation} 
where $\bm{W}_i^1, \bm{W}_i^2, \bm{W}_i^3$ are all $d_r \times d_k$ trainable matrices, $d_k=d_r/k$ is the dimension of each head vector, and $k$ is the number of head matrices. 
Then these $k$ heads are concatenated and linearly projected to the final feature set $\mathcal{M}$:
\begin{equation}
\small
\label{equ:multi_head}
    \mathcal{M} = \text{Concat}(\textbf{head}_1,...,\textbf{head}_k)\bm{W}_{C},
\end{equation}
where $\bm{W}_{C}$ is a ${d_r\times d_r}$ trainable matrix, $\mathcal{M}$ is the $N\times d_r$ feature set.

\noindent\textbf{\textsc{Function} Module.} 
It is designed to produce a single feature $\hat{\bm{v}}_F$ for generating function words like ``a'' or ``and''. The input of this module is a $d_c$ dimensional context vector $\bm{c}$ provided by the RNN, as the dashed line drawn in Figure~\ref{fig:pip}. We use $\bm{c}$ as the input because it contains rich language context knowledge of the partially generated captions, and such knowledge is suitable for generating function words, like ``a'' or ``and'', which require few visual knowledge. This module is formulated as:
\begin{equation} \label{equ:func_mod}
\small
\begin{aligned}
 \textbf{Input:} \quad  &\bm{c}, \\
 \textbf{Output:} \quad  &\hat{\bm{v}}_F=\text{LeakyReLU}(\text{FC}(\bm{c})), \\
\end{aligned}
\end{equation}
where $\hat{\bm{v}}_F$ is the $d_v$ dimensional output feature.

\subsection{Controller}
\label{sec:mod_col}
\begin{figure}[t]
\centering
\includegraphics[width=1\linewidth,trim = 5mm 5mm 5mm 5mm,clip]{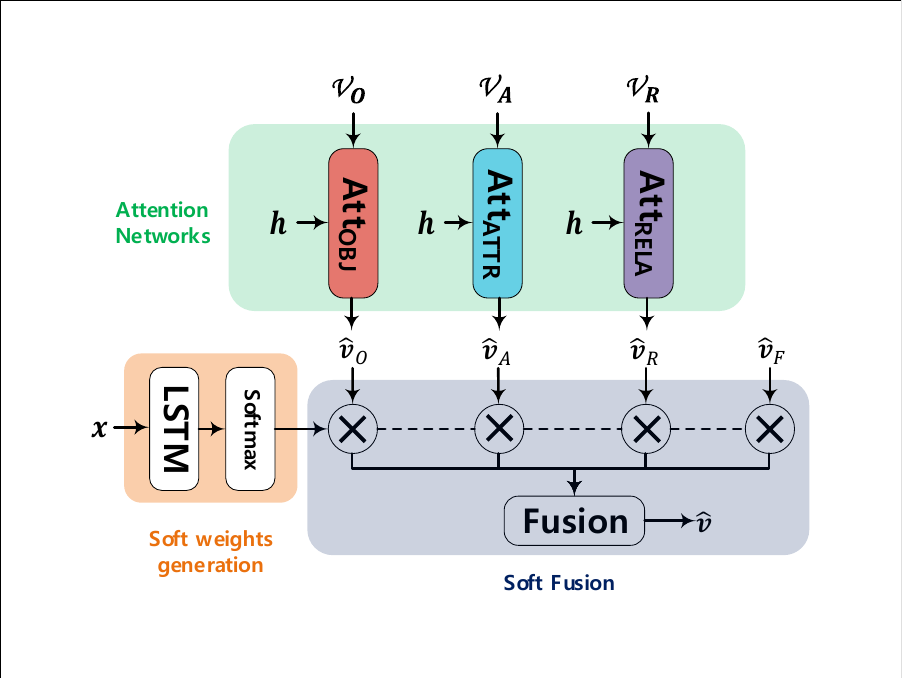}
   \caption{The detailed structure of our module controller. This controller will generate four soft weights by an LSTM for softly fusing attended features of four modules into a single fused feature $\hat{\bm{v}}$.}
\label{fig:cont}
\end{figure}
Figure~\ref{fig:cont} shows the detailed design of the module controller, which contains three attention networks, and one LSTM for soft weights generation. The output of this controller is a single fused feature vector $\hat{\bm{v}}$ which would be used for the next step reasoning by the followed RNN as in Figure~\ref{fig:pip}. Next, we describe our module controller.

\subsubsection{Soft Fusion}
\label{sec:soft_fuse}
Yet, it is still an open question on how to define a complete set of neural modules for visual reasoning~\cite{yu2018mattnet,andreas2016neural}. However, we believe that a combination of simple neural modules can approximate to accomplish a variety of complex tasks~\cite{hu2018explainable}. Before the soft fusion, three additive attention networks are used to respectively transform feature sets output from three visual modules into three more informative features:
\begin{equation} \label{equ:att_nets}
\small
\begin{aligned}
 \textbf{Object Attention:} \quad  &\hat{\bm{v}}_O = \text{Att}_{Obj}(\mathcal{V}_O,\bm{h}), \\
 \textbf{Attribute Attention:} \quad  &\hat{\bm{v}}_A = \text{Att}_{Attr}(\mathcal{V}_A,\bm{h}), \\
 \textbf{Relation Attention:} \quad  &\hat{\bm{v}}_R = \text{Att}_{Rela}(\mathcal{V}_R,\bm{h}),\\
\end{aligned}
\end{equation}
where $\hat{\bm{v}}_O$, $\hat{\bm{v}}_A$, and $\hat{\bm{v}}_R$ are the $d_v$ dimensional transformed features of $\mathcal{V}_O$, $\mathcal{V}_A$, and $\mathcal{V}_R$ produced by three visual modules (cf. Section~\ref{subsec:modules}), respectively; $\bm{h}$ is the $d_c$ dimensional query vector produced by an LSTM (specified in Section~\ref{sec:train_infer}); and the three attention networks own the same structure as that in~\cite{anderson2018bottom} while the parameters are not shared.

After getting the three transformed features, $\hat{\bm{v}}_O$, $\hat{\bm{v}}_A$, and $\hat{\bm{v}}_R$ from Eq.~\eqref{equ:att_nets} and the output $\hat{\bm{v}}_F$ from \textsc{function} module, the controller generates four soft weights for them. The process of generating soft weights is formulated as:
\begin{equation} \label{equ:soft_fuse}
\small
\begin{aligned}
 \textbf{Input:} \quad  &\bm{x}=\text{Concat}(\hat{\bm{v}}_O,\hat{\bm{v}}_A,\hat{\bm{v}}_R,\bm{c}), \\
 \textbf{Soft Vector:} \quad  &\bm{w} =\text{Softmax}(\text{LSTM}(\bm{x})), \\
 \textbf{Output:} \quad  &\hat{\bm{v}}=\text{Concat}(w_O \hat{\bm{v}}_O,w_A \hat{\bm{v}}_A,w_R \hat{\bm{v}}_R,w_F \hat{\bm{v}}_F), \\
\end{aligned}
\end{equation}
where the input $\bm{x}$ is the concatenation of three visual embedding vectors and the context vector accumulated in the RNN used in Eq.\eqref{equ:func_mod}; $\bm{w}=\{w_O,w_A,w_R,w_F\}$ is a four-dimensional soft attention vector; and the output vector $\hat{\bm{v}}$ will be fed into the RNN for the subsequent language decoding.

We use $\bm{x}$ for generating soft weights because both visual clues ($\hat{\bm{v}}_O,\hat{\bm{v}}_A,\hat{\bm{v}}_R$) and the language context knowledge $\bm{c}$ of partially generated captions are all indispensable for achieving satisfied module collocation. Also, since the layouts of modules at a new time step are highly related to the previous ones, an LSTM is applied here to accumulate such knowledge for generating new soft weights.

\subsubsection{Multi-Step Reasoning}
\label{sec:multi_res}
Different from many sentence-provided visual tasks like VQA where approximately perfect module layout can be parsed by the fully observed sentences, our module layout is still noisy because only partially observed sentences are available. To robustify the visual reasoning, we repeat the soft fusion and language decoding $M$ times as in~\cite{vaswani2017attention,Peters:2018,kim2018bilinear}. In this way, the generated captions are usually more relevant to the images by observing more visual clues. For example, as the experiment results shown in Section~\ref{sec:experiment}, when multi-step reasoning is implemented, more accurate quantifiers are generated because the visual patterns of the objects with the same category can be accumulated. In addition, residual connections (cf. Figure~\ref{fig:pip}) are used for directly transferring knowledge from lower layers to higher ones when such knowledge is already sufficient for word generation.

\subsubsection{Linguistic Loss}
\label{sec:ling_loss}
For ensuring each module to learn the orthogonal and non-trival knowledge from the image, \eg, \textsc{object} module focuses more on object categories instead of visual attributes, even it owns the same structure as \textsc{attribute} module. We design a linguistic loss which is imposed on the module controller for regularizing the training by making the controller faithful to human expert knowledge on part-of-speech collocation. 

We build this loss by extracting the words' lexical categories (\eg, adjectives, nouns, or verbs) from ground-truth captions by the Part-Of-Speech Tagger tool~\cite{toutanova2000enriching}. According to these lexical categories, we assign each word a 4-dimensional one hot vector $\bm{w}^{*}$, indicating which module should be chosen for generating this word. In particular, we assign \textsc{object} module to nouns (NN like ``bus''), \textsc{attribute} module to adjectives (ADJ like ``green''), \textsc{relation} module to verbs (VB like ``drive''), prepositions (PREP like ``on'') and quantifiers (CD like ``three''), and \textsc{function} module to the other words (CC like ``and'').

By providing these expert-guided module layout $\bm{w}^{*}$, the cross-entropy value between $\bm{w}^{*}$ and soft weights $\bm{w}$ in Eq.\eqref{equ:soft_fuse} is imposed to train the module controller:
\begin{equation}
\small
\label{equ:ling_loss}
    L_{lin}=-\sum_{i=1}^4{w_i^{*} \log w_i}.
\end{equation}
Note that this linguistic loss is imposed on all the $M$ module controllers in the language decoder (cf. Section~\ref{sec:multi_res}).

\subsection{Training and Inference}
\label{sec:train_infer}
By assembling the neural modules, module controller, ResNet-101~\cite{he2016deep} as CNN, and the top-down LSTM~\cite{anderson2018bottom} as RNN, our CNM image captioner can be trained end-to-end.
More specifically, at time step $t$, the query vector $\bm{h}$ in Eq.~\eqref{equ:att_nets} is the output of the first LSTM of the top-down structure at the same time step, and the context vector $\bm{c}$ in Eq.~\eqref{equ:func_mod} and Eq.~\eqref{equ:soft_fuse} is the output of the second LSTM of the top-down structure at time step $t-1$. 

Given a ground-truth caption $\mathcal{S}^{*}=\{s_{1:T}^{*}\}$ with its extracted part-of-speech tags $\bm{w}^{*}$, we can end-to-end train our CNM by minimizing the linguistic loss proposed in Eq.~\eqref{equ:ling_loss} and the language loss between the generated captions and the ground-truth captions. Suppose that the probability of word $s$ predicted by the language decoder of our CNM model is $P(s)$, we can define the language loss $L_{lan}$ as the cross-entropy loss:
\begin{equation}
\small
     L_{lan}=L_{XE} = -\sum_{t=1}^T \log P(s_t^*),
\label{equ:equ_celoss}
\end{equation}
or the negative reinforcement learning (RL) based reward~\cite{rennie2017self}:
\begin{equation}
\small
    L_{lan}=L_{RL} = -\mathbb{E}_{s_{t}^s \sim P(s)}[r(s_{1:T}^s;s_{1:T}^*)],
    \label{equ:equ_rlloss}
\end{equation}
where $r$ is a sentence-level metric for the sampled sentence $\mathcal{S}^s=\{s_{1:T}^s\}$ and the ground-truth $\mathcal{S}^{*}=\{s_{1:T}^{*}\}$, \eg, the CIDEr-D~\cite{vedantam2015cider} metric. Given the linguistic loss and language loss, the total loss is:
\begin{equation}
\small
    L = L_{lan} + \lambda L_{lin},
\label{equ:total_loss}
\end{equation}
where $\lambda$ is a trade-off weight. When inference in language generation, we adopt the beam search strategy~\cite{rennie2017self} with a beam size of 5.

\section{Experiments}
\subsection{Datasets, Settings, and Metrics}
\noindent\textbf{MS-COCO~\cite{lin2014microsoft}.} 
This dataset provides one official split: 82,783, 40,504 and 40,775 images for training, validation and test respectively. The 3rd-party Karpathy split~\cite{karpathy2015deep} was also used for the off-line test, which has 113,287, 5,000, 5,000 images for training, validation and test respectively.

\noindent\textbf{Visual Genome~\cite{krishna2017visual} (VG).} We followed Up-Down~\cite{anderson2018bottom} to use object and attribute annotations provided by this dataset to pre-train CNN. We filtered this noisy dataset by keeping the labels which appear more than $2,000$ times in the training set. After filtering, 305 objects and 103 attributes remain. Importantly, since some images co-exist in both VG and COCO, we also filtered out the annotations of VG which also appear in COCO test set.

\noindent\textbf{Settings.}
The captions of COCO were addressed by the following steps: the texts were first tokenized on white spaces; all the letters were changed to lowercase; the words were removed if they appear less than 5 times; each caption was trimmed to a maximum of $16$ words. At last, the vocabulary included totally $10,369$ words.

In Eq.~\eqref{equ:obj_mod}, $d_r$ and $d_v$ were set to 2,048 and 1,000 respectively; and in Eq.~\eqref{equ:func_mod}, $d_c$ was set to 1,000. The number of head vectors $k$ in Eq.~\eqref{equ:multi_head} was 8. At training time, Adam optimizer~\cite{kingma2014adam} was used and the learning rate was initialized to $5e^{-4}$ and was decayed by $0.8$ for every $5$ epochs. The cross-entropy loss Eq.~\eqref{equ:equ_celoss} and the RL-based loss Eq.~\eqref{equ:equ_rlloss} were in turn used to train our CNM 35 epochs and 100 epochs respectively. The batch size was set to 100. In our experiments, we found that the performance is non-sensitive to $\lambda$ in Eq.~\eqref{equ:total_loss}. By default, we set the trade-off weight $\lambda=1$ and $\lambda=0.5$ when the cross-entropy loss and RL-based loss were used as language loss, respectively.

\noindent\textbf{Metrics.}
\label{sec:metrics}
Five standard metrics were applied for evaluating the performances of the proposed method: CIDEr-D~\cite{vedantam2015cider}, BLEU~\cite{papineni2002bleu}, METEOR\cite{banerjee2005meteor}, ROUGE~\cite{lin2004rouge}, and SPICE~\cite{anderson2016spice}.

\subsection{Ablative Studies}
\label{sec:experiment}
We conducted extensive ablations for CNM, including architecture and fewer training sentences. 

\noindent\textbf{Architecture.}
We will investigate the effectiveness of designed modules, soft module fusion, linguistic loss, and deeper decoder structure in terms of proposing research questions (\textbf{Q}) and empirical answers (\textbf{A}).

\noindent\textbf{Q1:} Will each module generate more accurate module-specific words, \eg, will \textsc{object} module generate more accurate nouns? We deployed a single visual module as the encoder and the top-down attention LSTM~\cite{anderson2018bottom} as the decoder. When \textsc{object}, \textsc{attribute}, and \textsc{relation} modules were used, the baselines are denoted as \textbf{Module/O}, \textbf{Module/A}, and \textbf{Module/R}, respectively. In particular, baseline Module/O is the upgraded version of Up-Down~\cite{anderson2018bottom}.

\noindent\textbf{Q2:} Will the qualities of the generated captions be improved when the modules are fused? We designed three strategies for fusing modules by using three kinds of fusion weights. Specifically, when we set all the fusion weights as 1, the baseline is called \textbf{Col/1}; when soft fusion weights were used, the baseline is called \textbf{Col/S}; and when Gumbel-Softmax layer~\cite{jang2016categorical} was used for hard selection, the baseline is called \textbf{Col/H}. 

\noindent\textbf{Q3:} Will the expert knowledge of part-of-speech collocations provided by the linguistic loss benefit the model? We added the linguistic loss to baselines Col/H and Col/S to get baselines \textbf{Col/S+L} and \textbf{Col/H+L}, respectively. Noteworthy, linguistic loss can not be used to Col/1 since we do not need module controller here. 

\noindent\textbf{Q4:} Will better captions be generated when a deeper language decoder is implemented? We stacked the language decoder of baseline Col/S+L $M$ times to get baseline \textbf{CNM\#M}. Also, we designed \textbf{Module/O\#M} by stacking $M$ times of the top-down LSTM of baseline Module/O to check whether the performances can be improved when only the deeper decoder is used. 

\begin{figure}[t]
\centering
\includegraphics[width=1\linewidth,clip]{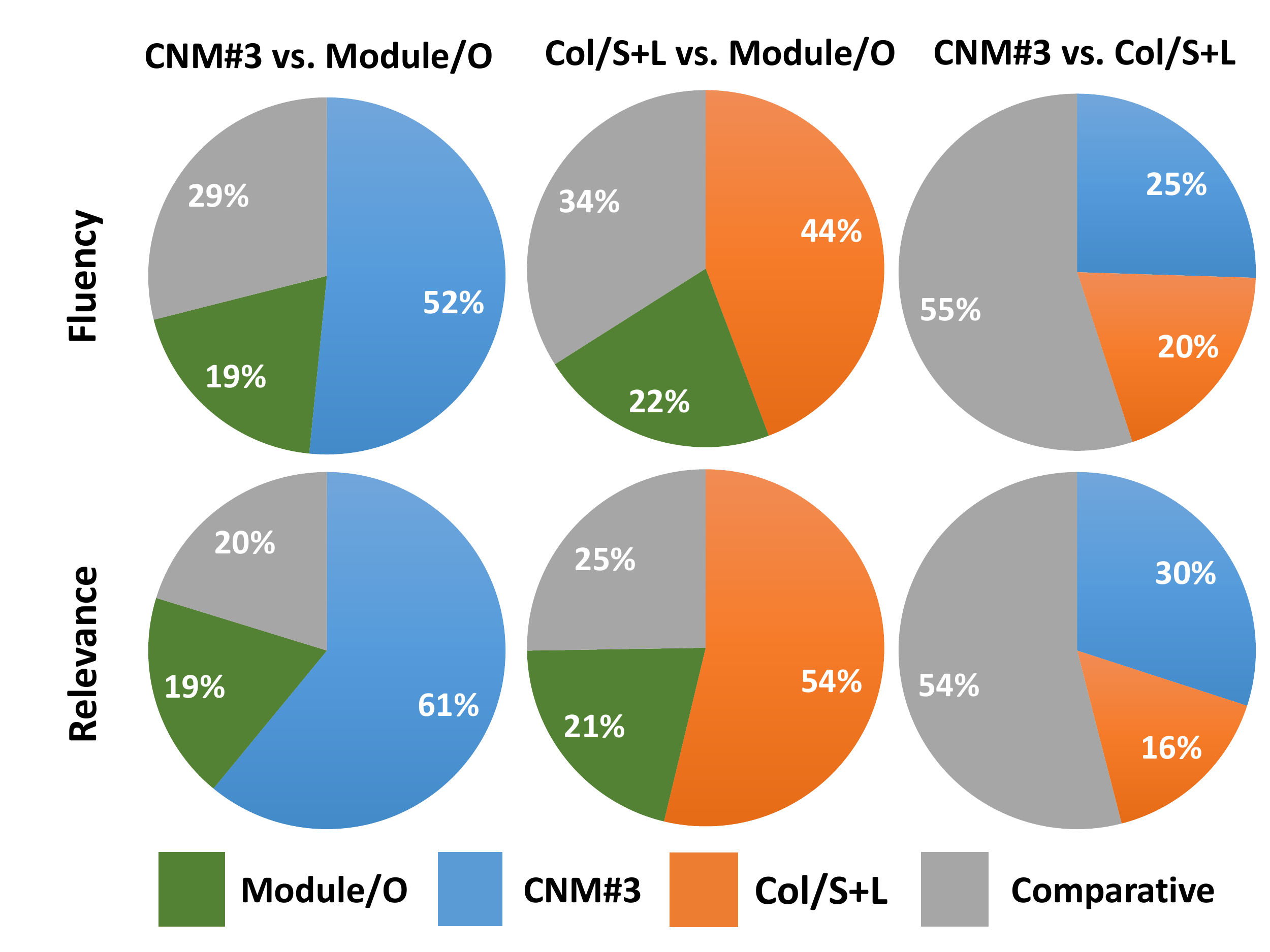}
  \caption{The pie charts each comparing the two methods in human evaluation.}
\label{fig:he}
\end{figure}

\noindent\textbf{Evaluation Metrics.}
For comprehensively validating the effectiveness of our CNM, we not only computed five standard metrics (cf. Section~\ref{sec:metrics}), but also conducted human evaluation and calculated the recalls of five part-of-speech words. Specifically, we invited 20 workers for human evaluation. We exhibited 100 images sampled from the test set for each worker and asked them to pairwisely compare the captions generated from three models: Module/O, Col/S+L, and CNM\#3. The captions are compared from two aspects: 1) \textbf{the fluency and descriptiveness of the generated captions} (the top three pie charts in Figure~\ref{fig:he});  2) \textbf{the relevance of the generated captions to images} (the bottom three pie charts in Figure~\ref{fig:he}). For calculating the recalls of five part-of-speech words, we counted the ratio of the words in predicted captions to the words in ground-truth captions. Such results are reported in Table~\ref{tab:statistic}.

\begin{table}[t]
\begin{center}
\caption{The performances of various methods on Karpathy split. The metrics: B@N, M, R, C, and S denote BLEU@N, METEOR, ROUGE-L, CIDEr-D, and SPICE, respectively.}
\label{table:tab_kap}
\scalebox{0.78}{
\begin{tabular}{l c c c c c c}
		\hline
		   Models   & B@1 & B@4 & M & R &  C & S\\ \hline
           SCST~\cite{rennie2017self}       & $-$  & $34.2$ & $26.7$ & $55.7$ & $114.0$ & $-$   \\ 
           LSTM-A~\cite{yao2017boosting}   & $78.6$  & $35.5$ & $27.3$ & $56.8$ & $118.3$ & $20.8$   \\ 
           StackCap~\cite{gu2017stack}    & $78.6$  & $36.1$ & $27.4$ & $-$ & $120.4$ & $-$   \\ 
           Up-Down~\cite{anderson2018bottom}  & $79.8$ & $36.3$ & $27.7$ & $56.9$ & $120.1$ & $21.4$ \\ 
           RFNet~\cite{jiang2018recurrent}  & $80.4$ & $37.9$ & $28.3$ & $58.3$ & $125.7$ & $21.7$ \\ 
           CAVP~\cite{liu2018context}  & $-$& $38.6$ & $28.3$ & $58.5$ & $126.3$ & $21.6$ \\
           SGAE~\cite{yang2018auto} & $\bm{80.8}$ & $38.4$ & $\bm{28.4}$ & $58.6$ & $127.8$ & $\bm{22.1}$ \\ \hline 
           Module/O & $79.6$ & $37.5$ & $27.7$ & $57.5$ & $123.1$ & $21.0$\\
           Module/A & $79.4$ & $37.3$ & $27.4$ & $57.1$ & $121.9$ & $20.9$\\
           Module/R & $79.7$ & $37.9$ & $27.8$ & $57.8$ & $123.8$ & $21.2$\\
           Module/O\#3 &$79.9$ & $38.0$ & $27.9$ & $57.5$ & $124.3$ & $21.3$\\
           Col/1 & $80.2$ & $38.2$ & $27.9$ & $58.1$ & $125.3$ & $ 21.3$\\
           Col/H & $80.1$ & $38.1$ & $27.8$ & $58.1$ & $124.7$ & $ 21.2$\\
           Col/H+L & $80.2$ & $38.3$ & $27.9$ & $58.4$ & $125.4$ & $ 21.4$\\
           Col/S & $80.2$ & $38.2$ & $28.0$ & $58.4$ & $125.7$ & $21.4$\\
           Col/S+L (CNM\#1) & $80.3$ & $38.5$ & $28.2$ & $58.6$ & $126.4$ & $21.5$\\
           CNM\#2 & $80.5$ & $38.5$ & $28.2$ & $58.7$ & $127.0$ & $21.7$\\
           CNM\#3 & $80.6$ & $38.7$ & $\bm{28.4}$ & $58.7$ & $127.4$ & $21.8$\\ 
           CNM\#3+SGAE & $\bm{80.8}$ & $\bm{38.9}$ & $\bm{28.4}$ & $\bm{58.8}$ & $\bm{127.9}$ & $22.0$ \\ \hline 
           
\end{tabular}
}
\end{center}
\vspace{-0.2in}
\end{table}
\begin{table}[t]
\begin{center}
\caption{The recalls (\%) of five part-of-speech words.}
\label{tab:statistic}
\scalebox{0.78}{
\begin{tabular}{l c c c c c}
		\hline
		   Models   & nouns & adjectives & verbs & prepositions & quantifiers \\ \hline
           Module/A & $42.4$ & $12.4$ & $20.2$ & $41.7$ & $14.3$ \\
           Module/O & $44.5$ & $11.5$ & $21.8$ & $42.6$ & $17.1$ \\ 
           Module/R & $44.3$ & $11.3$ & $22.8$ & $43.5$ & $22.3$ \\
           Col/S & $45.2$ & $13.1$ & $23.1$ & $43.6$ & $24.1$ \\
           Col/S+L & $45.9$ & $14.3$ & $23.5$ & $43.9$ & $25.4$ \\
           CNM\#3 & $\bm{47.3}$ & $\bm{16.1}$ & $\bm{24.3}$ & $\bm{44.8}$ & $\bm{30.5}$ \\ \hline
\end{tabular}
}
\end{center}
\vspace{-0.2in}
\end{table}

\noindent\textbf{A1.} 
From Table~\ref{tab:statistic}, we can observe that each single module prefers to generate more accurate module-specific words, \eg, the recall of nouns generated by Module/O is much higher than Module/A. Such observation validates that each module can indeed learn the knowledge of the corresponding module-specific words.

\noindent\textbf{A2.} 
As shown in Table~\ref{table:tab_kap}, when modules are fused, the performances can be improved. Also, by comparing Col/1, Col/S, and Col/H, we can find that Col/S achieves the highest performance. This is reasonable since compared with Col/1, Col/S can make word generation ground to the specific module. Compared with Col/H, Col/S can exploit more knowledge from all the modules when the modules are not correctly collocated. 

\noindent\textbf{A3.} 
As shown in Table~\ref{table:tab_kap} and~\ref{tab:statistic}, we can find that the performances of Col/S+L are better than Col/S. Such observations validate that the expert supervision can indeed benefit the caption generation. In addition, from the results shown in Figure~\ref{fig:he}, we can find that when soft module fusion and linguistic loss are deployed, the generated captions has higher qualities evaluated by humans.

\noindent\textbf{A4.} 
By inspecting the standard evaluation scores in Table~\ref{table:tab_kap}, the recalls of words in Table~\ref{tab:statistic}, and the human evaluations in Figure~\ref{fig:he}, we can find that when a deeper decoder is used, \eg, CNM\#3 vs. CNM\#1, the qualities of the generated captions can be improved. Also, by comparing Module/O\#3 with CNM\#3, we can find that only using a deeper decoder is not enough for generating high qualities captions.

\begin{table}[t]
\begin{center}
\caption{The CIDEr-D loss (CIDEr-D) of using fewer training sentences.}
\label{table:tab_fewer}
\scalebox{0.65}{
\begin{tabular}{l c c c c c}
		\hline
		   X   & 5 & 4 & 3 & 2 & 1 \\ \hline
           CNM\&X & $0 (127.4) $ & $\bm{0.4} (127.0)$ & $\bm{1.2}(126.2)$ & $\bm{2.3}(125.1)$ & $\bm{3.6}(123.8)$ \\
           Module-O\&X & $0 (123.1)$ & $0.9 (122.2)$ & $2.3 (120.8)$ & $4.1 (119.0)$ & $6.8 (116.3)$ \\ \hline
\end{tabular}
}
\end{center}
\vspace{-0.2in}
\end{table}

\noindent\textbf{Fewer Training Samples.}
To test the robustness of our CNM in the situation where only fewer training sentences are available (cf. Section~\ref{sec:intro}), we randomly assigned $X$ sentences among all the annotated captions to one image for training models CNM\#3 and Module-O to get baselines \textbf{CNM\&X} and \textbf{Module-O\&X}. The results are reported in Table~\ref{table:tab_fewer}, where the values mean the losses of CIDEr-D compared with the model trained by all sentences, and the values in the bracket are the CIDEr-D scores.

\noindent\textbf{Results and Analysis.}
From Table~\ref{table:tab_fewer}, we can find that both two models will be damaged if fewer training sentences are provided. Interestingly, we can observe that our CNM can halve the performance loss compared to Module/O. Such observations suggest that our CNM is more robust when fewer training samples are provided, compared with the traditional attention-based method.

\subsection{Comparisons with State-of-The-Arts}
\label{sec:sota}
\begin{figure}[t]
\centering
\includegraphics[width=1\linewidth,trim = 5mm 5mm 5mm 5mm,clip]{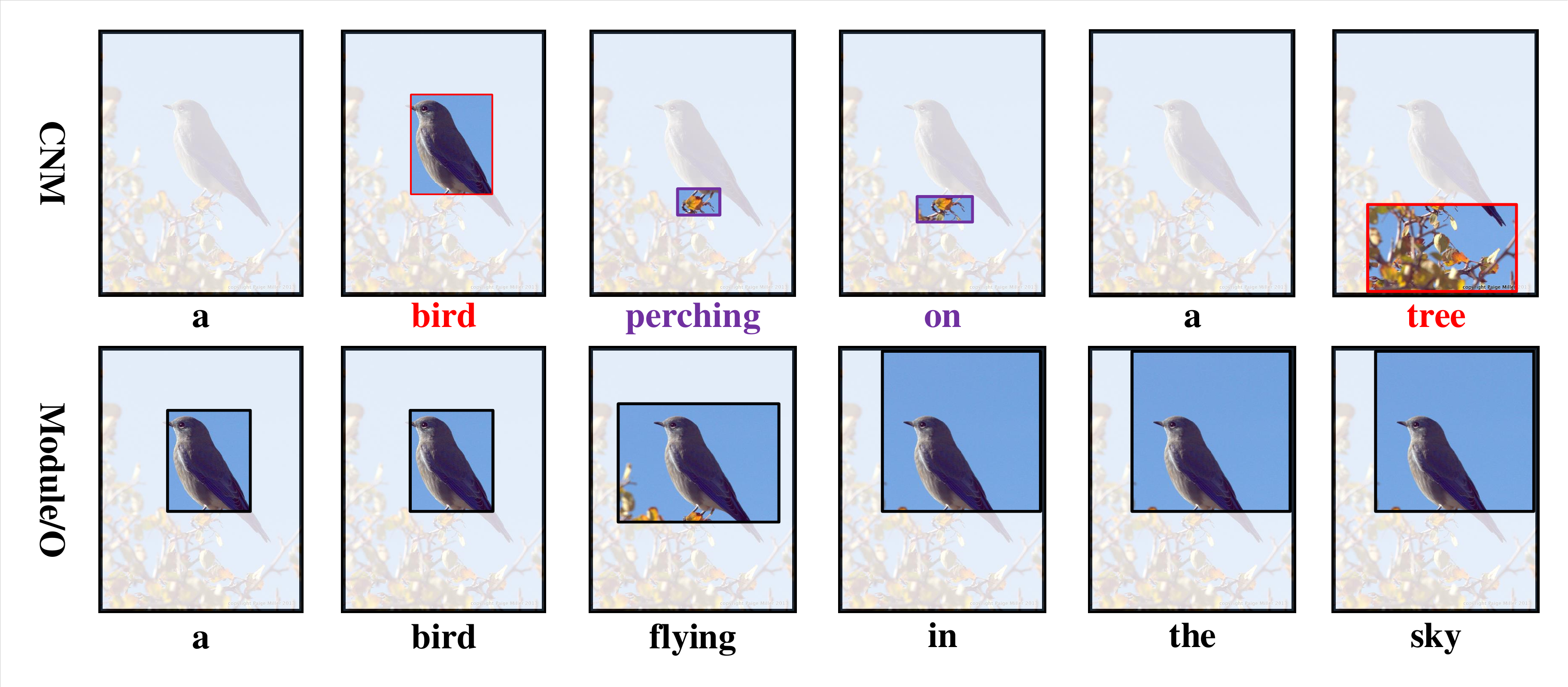}
  \caption{The visualizations of the caption generation process of two methods: CNM\#3 and Module/O. Different colours refer to different modules, \ie, red for \textsc{object} module, purple for \textsc{relation} module, and black for \textsc{function} module. For simplicity, we only visualize the module layout generated by the last module controller of the deeper decoder.}
\label{fig:demo}
\vspace{-0.2in}
\end{figure}

\noindent\textbf{Comparing Methods.} Though various captioning models are developed in recent years, for fair comparisons, we only compared our CNM with some encoder-decoder methods due to their superior performances. Specifically, we compared our method with \textbf{SCST}~\cite{rennie2017self}, \textbf{StackCap}~\cite{gu2017stack}, \textbf{Up-Down}~\cite{anderson2018bottom}, \textbf{LSTM-A}~\cite{yao2017boosting}, \textbf{NBT}~\cite{lu2018neural}, \textbf{CAVP}~\cite{liu2018context}, \textbf{RFNet}~\cite{jiang2018recurrent}, and \textbf{SGAE}~\cite{yang2018auto}. Among these methods, Up-Down and NBT are specific cases of our CNM where only \textsc{object} modules are deployed. All of StackCap, CAVP, and RFNet use wider encoders or deeper decoders, while they do not design different modules. In addition, we also equipped our CNM a dictionary preserving language bias as in SGAE~\cite{yang2018auto}, and this model is denoted as ~\textbf{CNM+SGAE}.

\noindent\textbf{Results.}
Table~\ref{table:tab_xe} and~\ref{table:tab_kap} show the performances of various methods trained by cross-entropy loss and RL-based loss, respectively. We can see that our single model CNM+SGAE in Table~\ref{table:tab_kap} achieves a new state-of-the-art CIDEr-D score. Specifically, by deploying four compact modules, soft module fusion strategy, and linguistic loss, our CNM can obviously outperform the models, \eg, StackCap, CAVP, and RFNet, which also use deeper decoders or wider encoders. When the dictionary preserving language bias is learned as in SGAE, even the query embeddings do not contain high-level semantic knowledge created by graph convolution network as SGAE, our CNM+SGAE also achieve better performances than SGAE. From the results of the online test reported in Table~\ref{table:tab_online}, we can find that our single model has competitive performances and can achieve the highest CIDEr-D c40 score. In addition, Figure~\ref{fig:demo} shows the visualizations of the captioning process of our CNM and Module/O (the upgraded version of Up-Down). From this figure, we can observe that our CNM can generate more relevant description ``bird perch'' and less overfitted to dataset bias of high co-occurrence word combination ``bird fly''.

\begin{table}[t]
\begin{center}
\caption{The performances of various methods on MS-COCO Karpathy split trained by cross-entropy loss.}
\label{table:tab_xe}
\scalebox{0.78}{
\begin{tabular}{l c c c c c c}
		\hline
		   Models   & B@1 & B@4 & M & R &  C & S\\ \hline
           SCST~\cite{rennie2017self}       & $-$  & $30.0$ & $25.9$ & $53.4$ & $99.4$ & $-$   \\ 
           LSTM-A~\cite{yao2017boosting}   & $73.4$  & $32.6$ & $25.4$ & $54.0$ & $100.2$ & $18.6$   \\ 
           StackCap~\cite{gu2017stack}    & $76.2$  & $35.2$ & $26.5$ & $-$ & $109.1$ & $-$   \\ 
           NBT~\cite{lu2018neural}  & $75.5$  & $34.7$ & $27.1$ & $-$ & $108.9$ & $20.1$   \\ 
           Up-Down~\cite{anderson2018bottom}  & $77.2$ & $36.2$ & $27.0$ & $56.4$ & $113.5$ & $20.3$ \\ 
           RFNet~\cite{jiang2018recurrent}  & $77.4$ & $37.0$ & $\bm{27.9}$ & $\bm{57.3}$ & $116.3$ & $\bm{20.8}$ \\ \hline
           Col/S+L (CNM\#1) & $77.3$ & $36.5$ & $27.6$ & $57.0$ & $116.4$ & $20.7$\\
           CNM\#3 & $\bm{77.6}$ & $\bm{37.1}$ & $\bm{27.9}$ & $\bm{57.3}$ & $\bm{116.6}$ & $\bm{20.8}$\\ \hline 
\end{tabular}
}
\end{center}
\vspace{-0.2in}
\end{table}

\begin{table}[t]
\begin{center}
\caption{The performances of various methods on the online MS-COCO test server.}
\label{table:tab_online}
\scalebox{0.68}{
\begin{tabular}{l c c c c c c c c c c c}
		\hline
		 Model  & \multicolumn{2}{c}{B@4} &\multicolumn{2}{c}{M} &\multicolumn{2}{c}{R-L} & \multicolumn{2}{c}{C-D}\\ \hline 
		 Metric  &   c5 & c40 & c5 & c40 & c5 & c40 & c5 & c40 \\ \hline
           SCST~\cite{rennie2017self}      & $35.2$ & $64.5$ & $27.0$ & $35.5$ & $56.3$ & $70.7$ & $114.7$ & $116.0$  \\
           LSTM-A~\cite{yao2017boosting}    & $35.6$ & $65.2$ & $27.0$ & $35.4$ & $56.4$ & $70.5$ & $116.0$ & $118.0$  \\ 
           StackCap~\cite{gu2017stack}     & $34.9$ & $64.6$ & $27.0$ & $35.6$ & $56.2$ & $70.6$ & $114.8$ & $118.3$   \\ 
           Up-Down~\cite{anderson2018bottom}    & $36.9$ & $68.5$ & $27.6$ & $36.7$ & $57.1$ & $72.4$ & $117.9$& $120.5$  \\ 
           CAVP~\cite{liu2018context}           & $37.9$ & $69.0$ & $28.1$ & $37.0$ & $58.2$ & $73.1$ & $121.6$& $123.8$    \\
           SGAE~\cite{yang2018auto}    &$37.8$ & $68.7$& $28.1$ & $37.0$ & $58.2$ & $73.1$ & $122.7$ & $125.5$\\ 
           CNM\#3    &$37.9$ & $68.4$& $28.1$ & $36.9$ & $58.3$ & $72.9$ & $123.0$ & $125.3$\\ 
           CNM+SGAE    &$\bm{38.4}$ & $\bm{69.3}$& $\bm{28.2}$ & $\bm{37.2}$ & $\bm{58.4}$ & $\bm{73.4}$ & $\bm{123.8}$ & $\bm{126.0}$\\ 
           \hline
\end{tabular}
}
\end{center}
\vspace{-0.2in}
\end{table}

\subsection{Limitations and Potentials}
\begin{figure}[t]
\centering
\includegraphics[width=1\linewidth,trim = 5mm 5mm 5mm 5mm,clip]{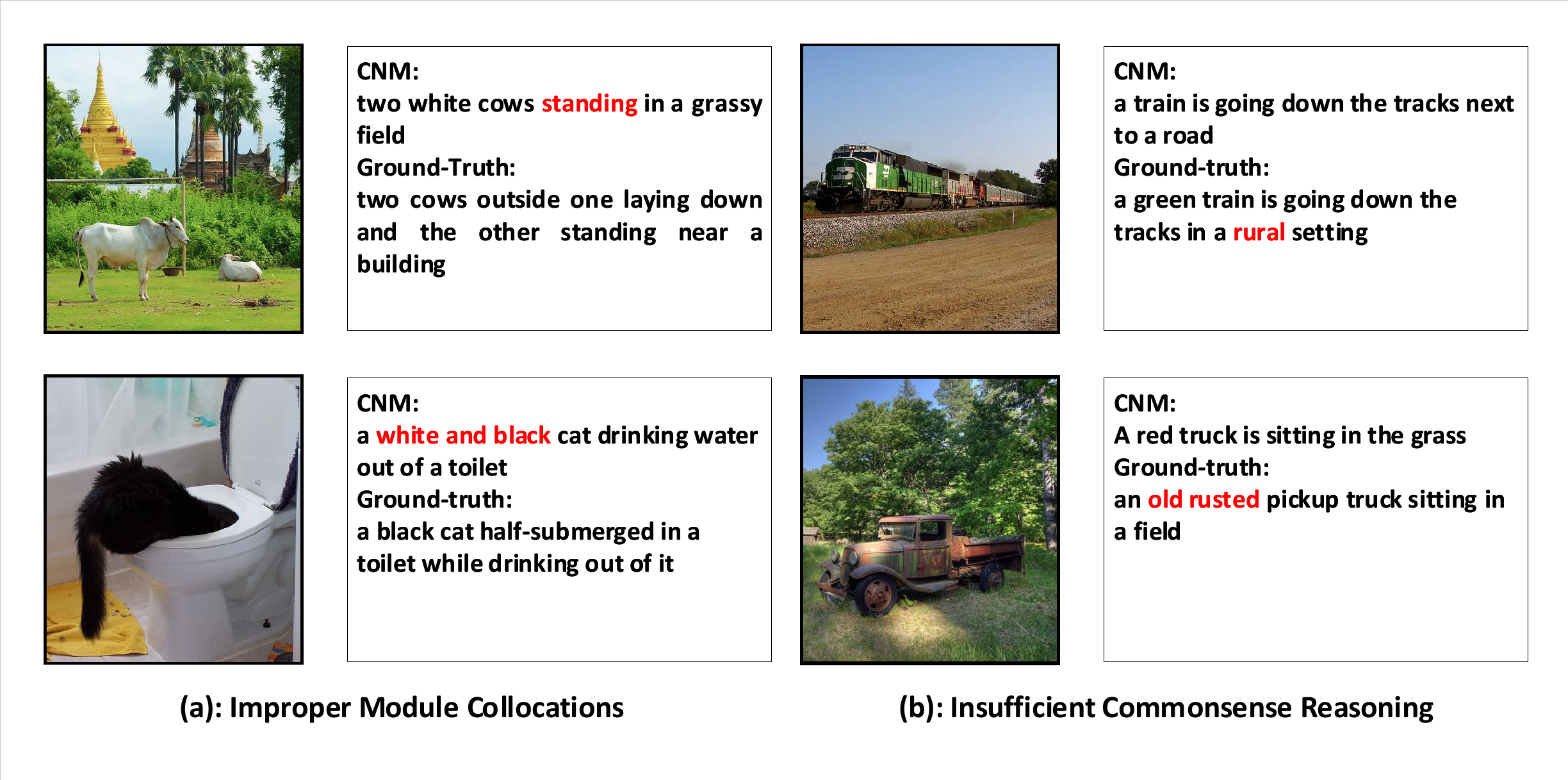}
  \caption{The limitations of our CNM model.}
  \vspace{-0.2in}
\label{fig:limit}
\end{figure}
Though we design three techniques, \eg, soft module fusion, linguistic loss, and multi-step reasoning for robustifying the module collocation, \textbf{improper module collocations} still exist since the sentence patterns are structured dynamically without a global ``oracle''. As a result, inaccurate description will be generated by given the improper module collocations. For example, as shown in Figure~\ref{fig:limit}a top, at time step 4, \textsc{relation} module is chosen inaccurately and the verb ``standing'' is generated, while two cows have different actions; in Figure~\ref{fig:limit}a bottom, at time step 3, it is more suitable to generate the noun ``toilet'', but \textsc{function} module is chosen and inaccurate description ``white and black cat'' is generated. For tackling this limitation, more advanced techniques like Reinforcement Learning could be exploited for guiding the module collocations.

Another limitation of our CNM is \textbf{insufficient commonsense reasoning}. Specifically, many adjectives which require commonsense reasoning can hardly be generated by our model, \eg, ``rural'', ``rusty'', or ``narrow'' are all commonsense adjectives. Figure~\ref{fig:limit}b gives two examples, where the words ``rusted'' and ``rural'' cannot be generated. One possible solution is to design a \textsc{reason} module where a memory network preserving the commonsense knowledge is exploited and then the context knowledge can be used as queries for reasoning. The model CNM+SGAE is one preliminary experiment designed for resolving such limitation. From Table~\ref{table:tab_kap}, we can see that the performance indeed improves. This may shed some light on using more sophisticated modules and commonsense reasoning strategies.

\section{Conclusions}
We proposed to imitate the humans inductive bias --- sentences are composed by structuring patterns first --- for image captioning. In particular, we presented a novel modular network method: learning to Collocate Neural Modules (CNM), which can generate captions by filling the contents into collocated modules. In this way, the caption generation is expected to be disentangled from dataset bias. We validated our CNM by extensive ablations and comparisons with state-of-the-art models
on MS-COCO. In addition, we discussed the model limitations and thus the corresponding potentials are  our future work.

{\small
\bibliographystyle{ieee}
\bibliography{egbib}
}

This supplementary document will further detail the following aspects in the main paper: A. Network Architecture, B. Details of Human Evaluations, C. More Qualitative Examples.
\section{Network Architecture}
Here, we introduce the detailed network architectures of all the components in our model, which includes four neural modules, a module controller, and decoders.

\subsection{Neural Modules}
In Section~3.1 of the main paper, we show how to use four neural modules to generate the orthogonal knowledge from the image. The detail structures of these four modules are respectively listed in the following tables: 1) \textsc{object} module in Table~\ref{table:tab_obj}, 2) \textsc{attribute} module in Table~\ref{table:tab_attr}, 3) \textsc{relation} module in Table~\ref{table:tab_rela}, and 4) \textsc{function} module in Table~\ref{table:tab_func}. In particular, the input vector $\bm{c}$ of \textsc{function} module in Table~\ref{table:tab_func} (1) is the output of an LSTM in the language decoder, and we will specify this context vector in Section~\ref{sec:supp_decoder}.

\begin{table*}[t]
\begin{center}
\caption{The details of \textsc{object} module.}
\label{table:tab_obj}
\begin{tabular}{|c|c|c|c|c|c|}
		\hline
		   \textbf{Index}&\textbf{Input}&\textbf{Operation}&\textbf{Output}&\textbf{Trainable Parameters}\\ \hline
		   (1)  &    -    &  RoI features  & $\mathcal{R}_O$ ($N \times 2,048$) & - \\ \hline
		   (2)  &    (1)    &  FC($\cdot$)    & $\mathcal{Z}_O$ ($N \times 1,000$) & FC($2,048 \rightarrow 1,000$ ) \\ \hline
		   (3)  &    (2)    &  Leaky ReLU  &  $\mathcal{V}_O$ ($N \times 1,000$) & - \\ \hline
\end{tabular}
\end{center}
\end{table*}

\begin{table*}[t]
\begin{center}
\caption{The details of \textsc{attribute} module.}
\label{table:tab_attr}
\begin{tabular}{|c|c|c|c|c|c|}
		\hline
		   \textbf{Index}&\textbf{Input}&\textbf{Operation}&\textbf{Output}&\textbf{Trainable Parameters}\\ \hline
		   (1)  &    -    &  RoI features  & $\mathcal{R}_A$ ($N \times 2,048$) & - \\ \hline
		   (2)  &    (1)    &  FC($\cdot$)   & $\mathcal{Z}_A$ ($N \times 1,000$) & FC($2,048 \rightarrow 1,000$ ) \\ \hline
		   (3)  &    (2)    &  Leaky ReLU  &  $\mathcal{V}_A$ ($N \times 1,000$) & - \\ \hline
\end{tabular}
\end{center}
\end{table*}

\begin{table*}[t]
\begin{center}
\caption{The details of \textsc{relation} module.}
\label{table:tab_rela}
\begin{tabular}{|c|c|c|c|c|c|}
		\hline
		   \textbf{Index}&\textbf{Input}&\textbf{Operation}&\textbf{Output}&\textbf{Trainable Parameters}\\ \hline
		   (1)  &    -    &  RoI features  & $\mathcal{R}_O$ ($N \times 2,048$) & - \\ \hline
		   (2)  &    (1)    &  
		   \begin{tabular}{c}
		         multi-head \\
		         self-attention (Eq.(4))  
		   \end{tabular}
		   & $\textbf{head}_i$ ($N \times 256$) & 
		   \begin{tabular}{c}
		         $\bm{W}_i^1$ ($2,048 \times 256$) \\
		         $\bm{W}_i^2$ ($2,048 \times 256$) \\
		         $\bm{W}_i^3$ ($2,048 \times 256$) 
		   \end{tabular}
		   \\ \hline
		   (3)  &    (2)    &  multi-head vector (Eq.(5))  &  $\mathcal{M}$ ($N \times 2,048$) & $\bm{W}_C$($2,048 \times 2,048$) \\ \hline
		   (4)  &    (3)    &
		   \begin{tabular}{c}
		         feed-forward \\
		         FC$_2$(ReLU(FC$_1$($\cdot$)))\\
		   \end{tabular}
		   & $\mathcal{V}_R$ ($N \times 1,000$) & 
		   \begin{tabular}{c}
		         FC$_1$ ($2,048 \rightarrow 2,048$) \\
		         FC$_2$ ($2,048 \rightarrow 1,000$) \\
		   \end{tabular}
		   \\ \hline
\end{tabular}
\end{center}
\end{table*}

\begin{table*}[t]
\begin{center}
\caption{The details of \textsc{function} module.}
\label{table:tab_func}
\begin{tabular}{|c|c|c|c|c|c|}
		\hline
		   \textbf{Index}&\textbf{Input}&\textbf{Operation}&\textbf{Output}&\textbf{Trainable Parameters}\\ \hline
		   (1)  &    -    &  context vector  & $\bm{c}$ ($1,000$) & - \\ \hline
		   (2)  &    (1)    &  FC($\cdot$)   & $\bm{z}_F$ ($1,000$) & FC($1,000 \rightarrow 1,000$ ) \\ \hline
		   (3)  &    (2)    &  Leaky ReLU  &  $\hat{\bm{v}}_F$ ($1,000$) & - \\ \hline
\end{tabular}
\end{center}
\end{table*}

\subsection{Module Controller}
\label{sec:supp_mod_cont}
In Section~3.2.1 of the main paper, we discuss how to use module controller to softly fuse four vectors generated by attention networks and \textsc{function} module. The common structure of three attention networks used in Eq.(7) and the detail process of soft fusion in Eq.(8) are demonstrated in Table~\ref{table:tab_attnet} and~\ref{table:tab_cont}, respectively. Specifically, the hidden vector $\bm{h}$ in Table~\ref{table:tab_attnet} (2) and the context vector $\bm{c}$ in Table~\ref{table:tab_cont} (1) are the outputs of two different LSTMs in the language decoder, and both of them will be specified in Section~\ref{sec:supp_decoder}.

\begin{table*}[t]
\begin{center}
\caption{The details of the common structure of three attention networks.}
\label{table:tab_attnet}
\begin{tabular}{|c|c|c|c|c|c|}
		\hline
		   \textbf{Index}&\textbf{Input}&\textbf{Operation}&\textbf{Output}&\textbf{Trainable Parameters}\\ \hline
		   (1)  &    -    &  feature set  & $\mathcal{V}$ ($N \times 1,000$) & - \\ \hline
		   (2)  &    -    &  hidden vector  & $\bm{h}$ ($1,000$) & - \\ \hline
		   (3)  &    (2)    &
		    \begin{tabular}{c}
		         attention weights \\
		         $\bm{w}_a\tanh(\bm{W}_v\bm{v}_{n}+\bm{W}_h\bm{h})$\\
		   \end{tabular}
		     & $\bm{\alpha}$ ($N$) &
		   \begin{tabular}{c}
		         $\bm{w}_a$ (512), $\bm{W}_v$ ($512\times1,000$) \\
		         $\bm{W}_h$($512\times1,000$)
		   \end{tabular}
		   \\ \hline 
		   (4)  &    (3)     & Softmax & $\bm{\alpha}$ ($N$) & - \\ \hline
		   (5)  & (1),(4) & weighted sum $\bm{\alpha}^T\mathcal{V}$ & $\hat{\bm{v}}$ ($1,000$) & - \\ \hline 
\end{tabular}
\end{center}
\end{table*}

\begin{table*}[t]
\begin{center}
\caption{The details of soft fusion.}
\label{table:tab_cont}
\begin{tabular}{|c|c|c|c|c|c|}
		\hline
		   \textbf{Index}&\textbf{Input}&\textbf{Operation}&\textbf{Output}&\textbf{Trainable Parameters}\\ \hline
		   (1)  &    -    &  context vector  & $\bm{c}$ ($1,000$) & - \\ \hline
		   (2)  &    -    &  attended object feature  & $\hat{\bm{v}}_O$ ($1,000$) & - \\ \hline
		   (3)  &    -    &  attended attribute feature  & $\hat{\bm{v}}_A$ ($1,000$) & - \\ \hline
		   (4)  &    -    &  attended relation feature  & $\hat{\bm{v}}_R$ ($1,000$) & - \\ \hline
		   (5)  &    -    &  function feature  & $\hat{\bm{v}}_F$ ($1,000$) & - \\ \hline
		   (6)  &    (1),(2),(3),(4)  &  Concatenate   & $\bm{x}$ ($4,000$) & - \\ \hline
		   (7)  & (6)  & LSTM$_C$ $(\bm{x};\bm{h}_{C}^{t-1})$  & $\bm{h}_{C}^{t}$ (1,000) &   LSTM$_C$ (4,000 $\rightarrow$ $1,000$)\\ \hline 
		   (8) & (7) & Softmax & $\bm{w}$ (4) & - \\ \hline
		   (9) & (2),(3),(4),(8) & $\hat{\bm{v}}=\text{Concat}(w_O \hat{\bm{v}}_O,w_A \hat{\bm{v}}_A,w_R \hat{\bm{v}}_R,w_F \hat{\bm{v}}_F$) & $\hat{\bm{v}}$ ($4,000$) & - \\ \hline
\end{tabular}
\end{center}
\end{table*}

\subsection{Language Decoder}
\label{sec:supp_decoder}
As discussed in Section~3.2 of the main paper, the whole language decoder is built by stacking $M$ single language decoders with a common structure while the parameters are different. We set the top-down LSTM~\cite{anderson2018bottom} as our single language decoder and its architecture is shown in Table~\ref{table:tab_dec}. Specifically, for the $m$-th decoder, the input $\bm{i}^{m-1}$ in Table ~\ref{table:tab_dec} (1) is the output of the $m-1$-th decoder. When $m=1$, this input is word embedding vector $\bm{W}_{\Sigma}\bm{s}_{t-1}$, where $\bm{W}_{\Sigma}$ is a trainable embedding matrix and $\bm{s}_{t-1}$ is the one-hot vector of the word generated at time step $t-1$. In Table~\ref{table:tab_dec} (2), the output of the second LSTM $\bm{h}_2^{t-1}$ at time step $t-1$ is used as the context vector $\bm{c}$ in Table~\ref{table:tab_func} (1) and Table~\ref{table:tab_cont} (1), and the output of the first LSTM $\bm{h}^{t}_1$ in Table~\ref{table:tab_dec} (11) is used as the hidden vector $\bm{h}$ in Table~\ref{table:tab_attnet} (2). After getting the output of the $M$-th language decoder $\bm{i}^M$, a fully connected layer and softmax activation are used for producing the word distribution $P(s)$ (cf. Section~3.3 of the main paper).

\begin{table*}[t]
\begin{center}
\caption{The details of the single language decoder.}
\label{table:tab_dec}
\begin{tabular}{|c|c|c|c|c|}
		\hline
		   \textbf{Index}&\textbf{Input}&\textbf{Operation}&\textbf{Output}&\textbf{Trainable Parameters}\\ \hline
		   (1)  & - & the output of the last decoder     &  $\bm{i}^{m-1}$ ($1,000$) & - \\ \hline
		   (2)  & - & the output of LSTM$_2^m$ at $t-1$ & $\bm{h}_2^{t-1}$ (1,000) & -  \\ \hline
		   (3)  & - & object feature set & $\mathcal{V}_O$ ($N \times 1,000$) & -  \\ \hline
		   (4)  & - & attribute feature set & $\mathcal{V}_A$ ($N \times 1,000$) & -  \\ \hline
		   (5)  & - & relation feature set & $\mathcal{V}_R$ ($N \times 1,000$) & -  \\ \hline
		   (6)  & - & function feature & $\hat{\bm{v}}_F$ ($N \times 1,000$) & -  \\ \hline
		   (7)  & (3)    & mean pooling & $\bar{\bm{v}}_O$ ($1,000$) & - \\ \hline 
		   (8)  & (4)    & mean pooling & $\bar{\bm{v}}_A$ ($1,000$) & - \\ \hline 
		   (9)  & (5)    & mean pooling & $\bar{\bm{v}}_R$ ($1,000$) & - \\ \hline 
		   (10)  & (1),(2),(7),(8),(9) & concatenate & $\bm{u}^t$ ($5,000$) & - \\ \hline
		   (11)  & (10)  & LSTM$_1^m$ $(\bm{u}^t;\bm{h}^{t-1}_1)$  & $\bm{h}^{t}_1$ ($1,000$) &  LSTM$_1^m$ ($5,000 \rightarrow 1,000$)  \\ \hline
		   (12)  & (3),(11)  & attention network (Table~\ref{table:tab_attnet})  & $\hat{\bm{v}}_O$ ($1,000$) & -  \\ \hline
		   (13)  & (4),(11)  & attention network (Table~\ref{table:tab_attnet})  & $\hat{\bm{v}}_A$ ($1,000$) & -  \\ \hline
		   (14)  & (5),(11)  & attention network (Table~\ref{table:tab_attnet})  & $\hat{\bm{v}}_R$ ($1,000$) & -  \\ \hline
		   (15)  & (2),(6),(12),(13),(14)  & soft fusion (Table~\ref{table:tab_cont})  & $\hat{\bm{v}}^t$ ($4,000$) & -  \\ \hline
		   (16) & (11),(15) & LSTM$_2^m$ $([\bm{h}^t_1,\hat{\bm{v}}^t];\bm{h}^{t-1}_2)$ & $\bm{h}^t_2$ ($1,000$) & LSTM$_2^m$ ($5,000 \rightarrow 1,000$) \\ \hline
		   (17) & (1),(16) & add & $\bm{i}^m$ ($1,000$) & - \\ \hline
\end{tabular}
\end{center}
\end{table*}

\section{Human Evaluation}
\begin{figure*}[t]
\centering
\includegraphics[width=1\linewidth,trim = 5mm 5mm 5mm 5mm,clip]{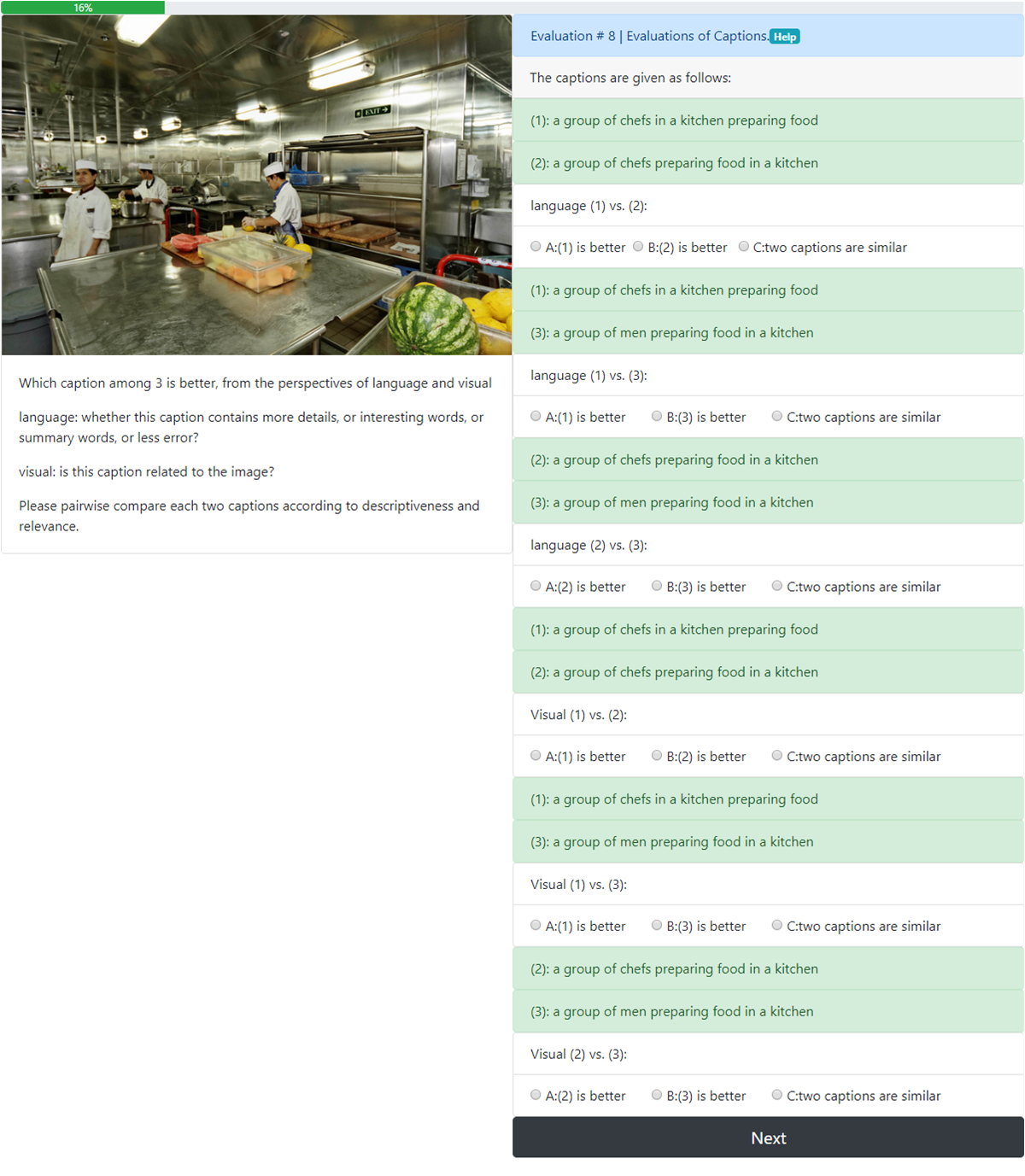}
  \caption{The evaluation interface for comparing captions generated by different models.}
  \vspace{-0.2in}
\label{fig:supp_he}
\end{figure*}

In the experiment (cf. Section~4.2 and Figure~5 of the main paper), we conducted human evaluation for better evaluating the qualities of the captions generated by different methods. In humane evaluation, the invited workers were required to compare the captions from two perspectives: 1) the fluency, \eg, less grammar error, and descriptiveness, \eg, more human-like descriptions, of the generated captions, and 2) the relevance of the generated captions to images. Figure~\ref{fig:supp_he} shows one example of the interface of our human evaluation.

\section{More Qualitative Examples}
\begin{figure*}[t]
\centering
\includegraphics[width=1\linewidth,trim = 5mm 5mm 5mm 5mm,clip]{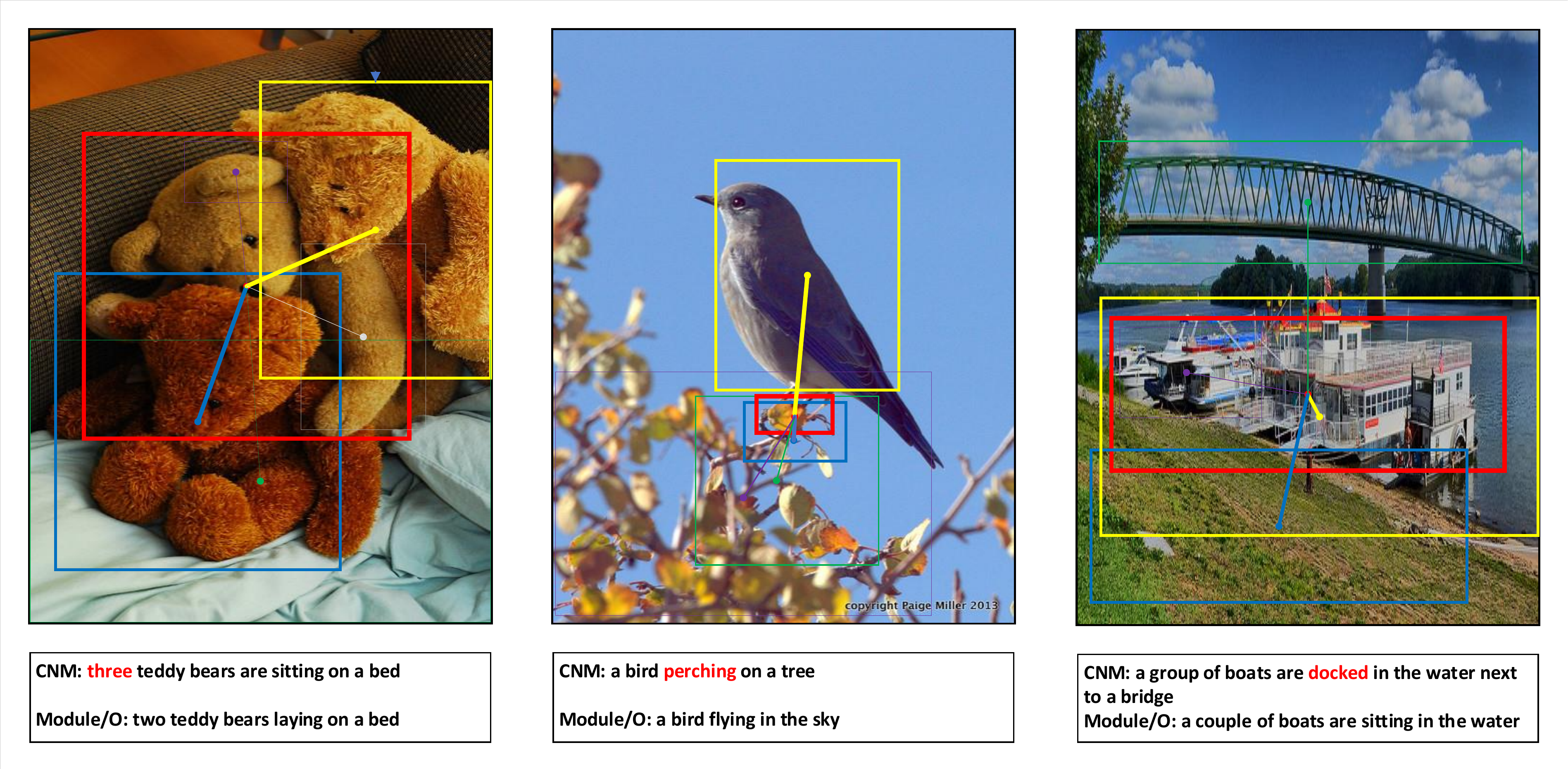}
  \caption{Three visualizations show how \textsc{relation} module generates relation specific words like quantifiers and verbs. The red box in each image is the attended image region (with the largest soft weight) when \textsc{relation} module generates a relation specific word. The thickness of lines connecting different boxes is determined by the soft attention weights computed by self-attention technique in Eq.(4). The thicker the line connecting two boxes is, the larger the soft weight between two bounding boxes is.}
\label{fig:supp_rela}
\end{figure*}

Figure~\ref{fig:supp_rela} exhibits three visualizations for explaining how \textsc{relation} module generates relation specific words. For example, in the middle figure, at the third time step, \textsc{relation} module focuses more on the ``paw'' part (red box) of one bird, and meantime the knowledge about ``bird'' (yellow box) and ``tree'' (blue box) is also incorporated to the ``paw'' part of the bird by multi-head self-attention technique (cf. Eq.(4) of the main paper). By exhaustively considering these visual clues, a more accurate action ``perch'' is generated.

\begin{figure*}[t]
\centering
\includegraphics[width=1\linewidth,trim = 5mm 5mm 5mm 5mm,clip]{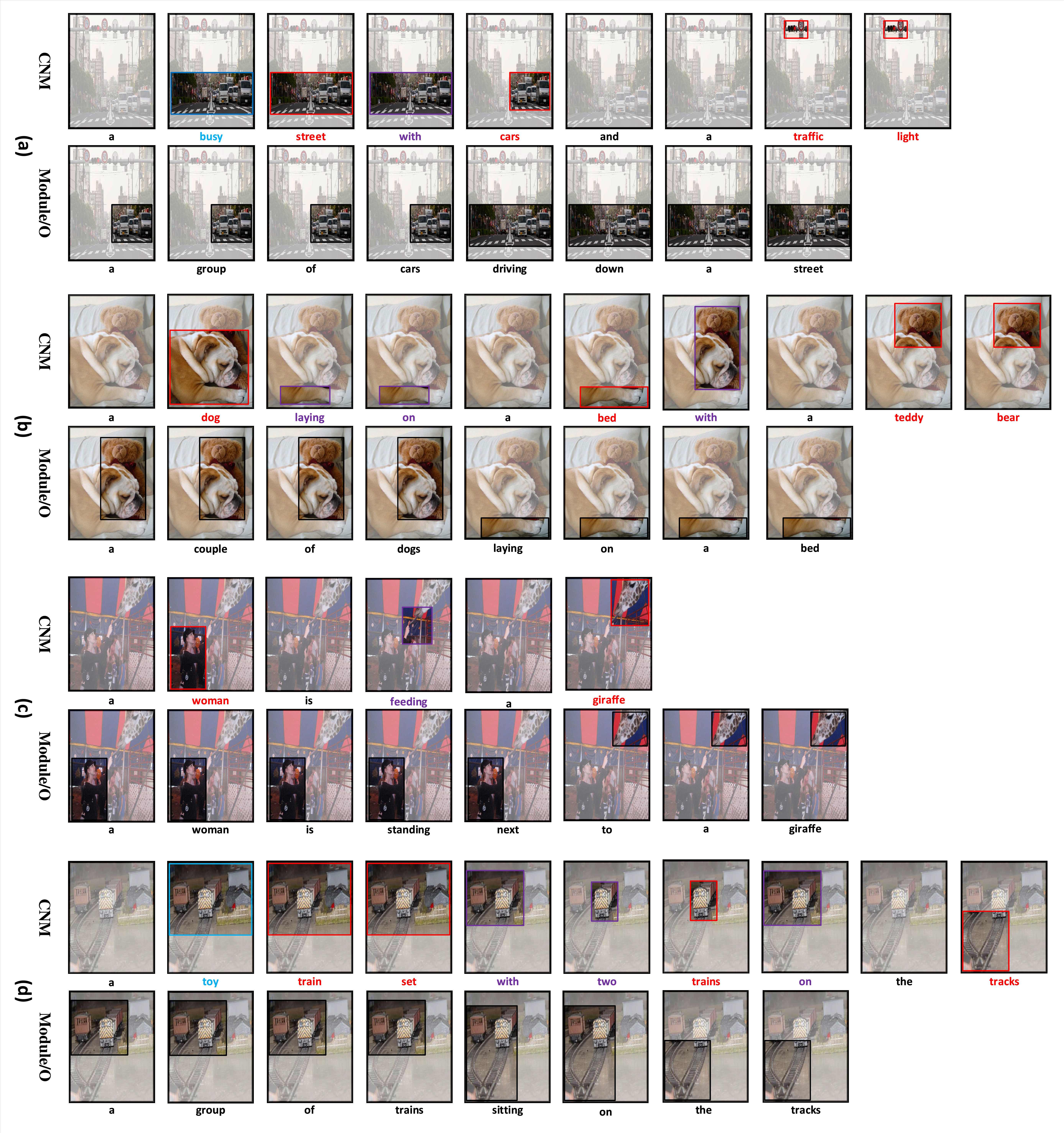}
  \caption{The visualizations of the caption generation process of two methods: CNM\#3 and Module/O. For CNM, different colours refer to different modules, \ie, blue for \textsc{attribute} module, red for \textsc{object} module, purple for \textsc{relation} module, and black for \textsc{function} module. For simplicity, we only visualize the module layout generated by the last module controller of the deeper decoder and only the image region with the largest soft weight is shown. For Module/O, only image region with the largest soft weight is visualized with black boundary.}
\label{fig:supp_demo}
\end{figure*}

Figure~\ref{fig:supp_demo} shows more comparisons between captions generated by CNM and Module/O. We can find that compared with Module/O, our CNM prefers to use some more accurate words to describe the appeared objects, attributes, and relations. For example, in Figure~\ref{fig:supp_demo} (a), the attribute ``busy'' can be assigned to ``street'', and in Figure~\ref{fig:supp_demo} (c), the action ``feed'' can be correctly generated. 

\end{document}